\def\year{2020}\relax
\documentclass[letterpaper]{article} 
\usepackage{aaai20}  
\usepackage{times}  
\usepackage{helvet} 
\usepackage{courier}  
\usepackage[hyphens]{url}  
\usepackage{graphicx} 
\usepackage{array}
\usepackage{multirow}
\usepackage{url}
\usepackage[switch]{lineno}
\usepackage{amsfonts,amssymb}

\usepackage{graphicx}  
\title{Viewpoint-Aware Loss with Angular Regularization for Person Re-Identification}

\author{
Zhihui Zhu\textsuperscript{\rm 1, \rm2} \thanks{Major work was done when the author worked in YouTu Lab.},
Xinyang Jiang\textsuperscript{\rm 1},
Feng Zheng\textsuperscript{\rm 3},
Xiaowei Guo \textsuperscript{\rm 2}, 
Feiyue Huang\textsuperscript{\rm 1}, \\
\Large \textbf{ 
Xing Sun\textsuperscript{\rm 1} \thanks{Corresponding Author: winfredsun@tencent.com},
Weishi Zheng \textsuperscript{\rm 2} \thanks{Corresponding Author: wszheng@ieee.org}}\\
\textsuperscript{\rm 1} Tencent YouTu Lab, Shanghai, China \\
\textsuperscript{\rm 2} Sun Yat-sen University, Guangzhou, China \\
\textsuperscript{\rm 3} Southern University of Science and Technology, Shenzhen, China \\
zhuzhh27@mail2.sysu.edu.cn, winfredsun@tencent.com, wszheng@ieee.org
}
\begin{document}

\maketitle
\begin{abstract}

Although great progress in supervised person re-identification (Re-ID) has been made recently, due to the viewpoint variation of a person, Re-ID remains a massive visual challenge. Most existing viewpoint-based person Re-ID methods project images from each viewpoint into separated and unrelated sub-feature spaces. They only model the identity-level distribution inside an individual viewpoint but ignore the underlying relationship between different viewpoints. To address this problem, we propose a novel approach, called \textit{Viewpoint-Aware Loss with Angular Regularization }(\textbf{VA-reID}). Instead of one subspace for each viewpoint, our method projects the feature from different viewpoints into a unified hypersphere and effectively models the feature distribution on both the identity-level and the viewpoint-level. In addition, rather than modeling different viewpoints as hard labels used for conventional viewpoint classification, we introduce viewpoint-aware adaptive label smoothing regularization (VALSR) that assigns the adaptive soft label to feature representation. VALSR can effectively solve the ambiguity of the viewpoint cluster label assignment. Extensive experiments on the Market1501 and DukeMTMC-reID datasets demonstrated that our method outperforms the state-of-the-art supervised Re-ID methods.

\end{abstract}

\section{Introduction}
\label{sec:da}
Person re-identification(Re-ID) which targets at recognizing pedestrians across non-overlapping camera views, is an important and challenging problem in visual surveillance analysis and has drawn increasing research attention \cite{zheng2015scalable,Zhao_2017_ICCV,Sun_2018_ECCV}. While Re-ID has gained considerable development in recent years, existing supervised person re-identification still faces some major visual appearance challenges, such as changes in viewpoint or poses, low resolution, illumination and etc.

\begin{figure}[t]
\centering
\includegraphics[width=0.9\columnwidth]{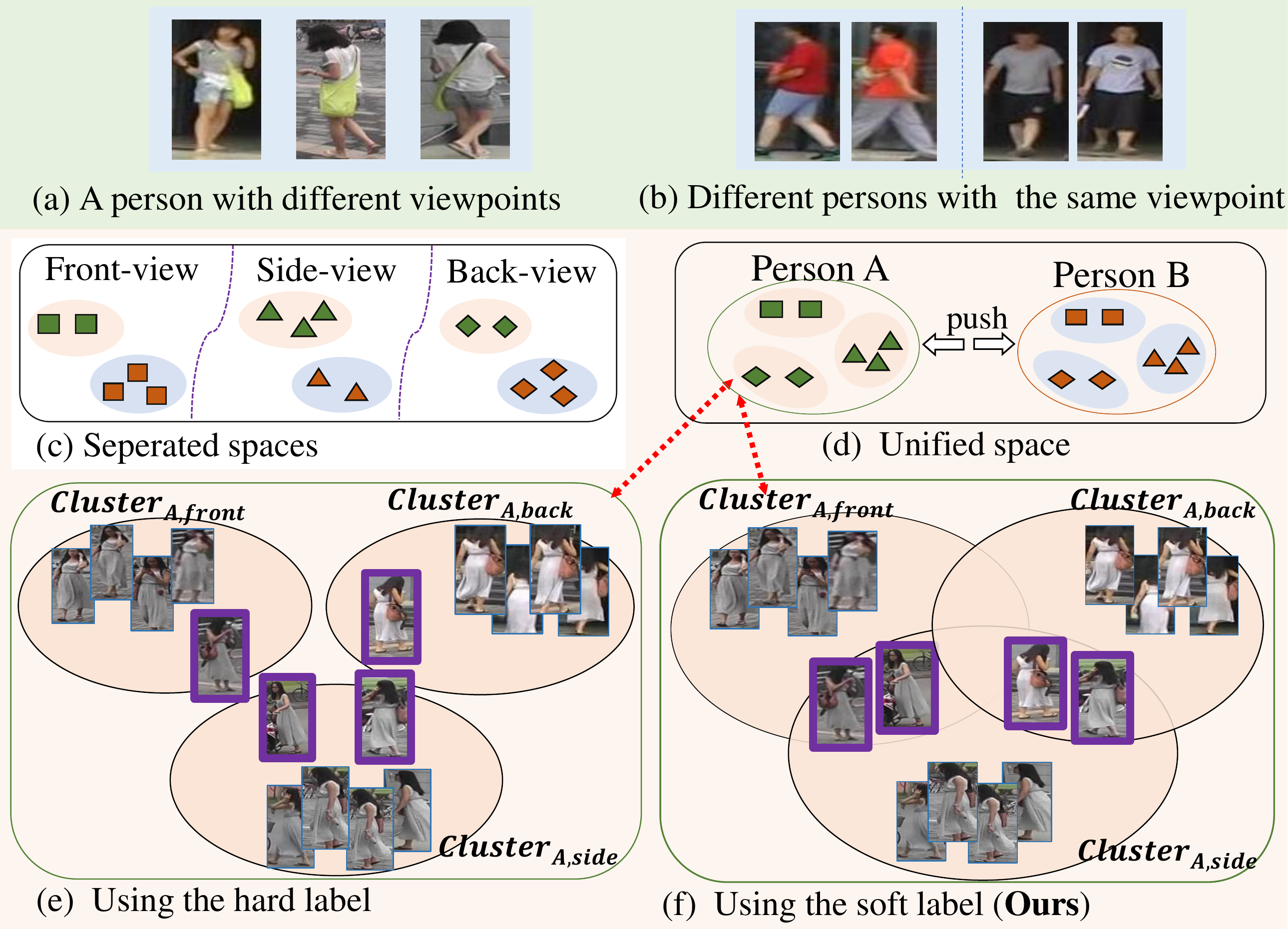} 
\caption{Comparisons of different feature learning methods. Our VA-reID method learns the unified space using the soft label instead of the hard label. Images with the thick purple border in the figure are the ambiguous viewpoint categories.}
\label{fig:intro}
\end{figure}

Among these challenges, in this work, we focus on the problem of viewpoint variations, which is one of the most important and difficult challenges in Re-ID research and practical application. In practice, due to the effect of viewpoint variation, images from different viewpoints of the same identity usually have massive visual appearance differences, and it may even be possible that some images of different identities from the same viewpoints are more similar in visual appearance than images of the same identity from different viewpoints. Some examples are shown in Figure \ref{fig:intro} (a)(b). This problem greatly limits the practical application of Re-ID.

A key problem for tackling viewpoint variations is to learn discriminative feature representations for body images with different viewpoints. However, there are some inadequacies of the existing viewpoint-based feature learning method \cite{ijcai2018-86,chen2018person,Qian_2018_ECCV,Sarfraz_2018_CVPR}. 1) They treat viewpoint learning and identity discrimination as two separate progresses, In such a case, it is not a principle way to learn optimal identity classification under various viewpoint variations. 2) They cast viewpoints of persons as hard labels, while in reality the viewpoint of person is ambiguous. As shown in Figure \ref{fig:intro}(c), these methods learn separate features for different viewpoints. For example, DVAML \cite{ijcai2018-86} learns two different feature subspaces for image pairs with similar and dissimilar viewpoints, OSCNN \cite{chen2018person} and PSE \cite{Sarfraz_2018_CVPR} learn a linear combination of different viewpoints' features. Actually, projecting the features into separated and unrelated subspaces only models the identity-level distribution within each viewpoint but may ignore the underlying relationship among different viewpoints. Thus, the relationship between the features from separate viewpoint subspaces cannot be directly learned, compromising the model's ability to match images of a person from different viewpoints.

To solve this problem, we propose a novel angular-based feature learning method that projects all features into a unified subspace and directly models the distribution of the features from different viewpoints. As shown in Figure \ref{fig:intro}(d)(f), the feature distribution is modeled at both the identity-level and viewpoint-level. At the identity-level, different identities are pushed away from each other to form identity-level clusters. At the viewpoint-level, the features in each identity cluster will further produce three viewpoint-level clusters (front, side, back), and a novel center regularization is used to pull the centers of these clusters closer to each other due to their visual similarity to the same identity.

In addition, we further consider the problem of the viewpoint cluster label assignment. While conventionally, each image will be assigned a hard viewpoint cluster label, we find that the viewpoint of some image samples are indeed ambiguous, as shown in Figure \ref{fig:intro}(e), and that a hard viewpoint cluster label assignment may mislead the learning. Therefore, we propose to relax the hard assignment problem and instead perform a soft label assignment, as shown in Figure \ref{fig:intro} (f). We propose a novel viewpoint-aware regularization method, called view aware label smoothing regularization (VALSR). VALSR replaces the common one-hot hard label with the adaptive soft label that changes adaptively according to similarity to the classification centers. Notice that we use the viewpoint label \textbf{only for training}.

In summary, this work develop a joint learning model for the identity and viewpoint discrimination learning, where in particular we introduce the soft multilabel to model viewpoint aware feature distribution for overcoming the ambiguity problem on viewpoint labelling and greatly solve the viewpoint variations. The main contributions include:
\begin{itemize}
    \item We put forward the idea of modeling the viewpoint distribution and identity distribution jointly rather than separately. For this purpose, we propose a novel solution called Viewpoint-Aware Loss with Angular Regularization, which effectively models the distribution on both identity-level and viewpoint-level, and specially we impose the center regularization to connect identity and viewpoint discrimination. The experimental results also demonstrate our advanced performance against related methods a lot.
    \item To overcome the ambiguity on viewpoint labelling, we develop viewpoint-aware adaptive label smoothing, which allows smooth transition between features of different discrete veiwpoints by assigning adaptive soft viewpoint label. The soft label will get self-adaption dynamically according to the prediction probability and have better performance against noise data and over-fitting
\end{itemize}

\section{Related Work}
\noindent \textbf{Viewpoint-Aware Person Re-identification.} 
Person Re-ID aims to recognize pedestrians across non-overlapping camera views. A recent survey \cite{leng2019survey} provides a detailed review and prospection of person Re-ID. Existing studies on viewpoint variation mainly lie in four aspects: pose-based methods \cite{Su_2017_ICCV,Sarfraz_2018_CVPR}, segmentation-based methods \cite{Kalayeh_2018_CVPR}, generation-based methods \cite{Zhou_2018_CVPR,Qian_2018_ECCV,Sun_2019_CVPR} and viewpoint-based methods \cite{ijcai2018-86}. Pose-based methods and segmentation-based methods are common practices to address this problem.  Pose-based methods usually take advantage of pose information to pay attention to human body parts while segmentation-based methods utilize human parsing information to obtain position information of human body parts. Then they can make align the body parts or extract the local feature of body parts. Generation-based methods usually use the generative model to generate images \cite{Qian_2018_ECCV} or generate feature of other viewpoints \cite{Zhou_2018_CVPR}. Recently PersonX \cite{Sun_2019_CVPR} utilizes a large-scale synthetic data engine to generate pedestrian images with arbitrary rotation angle, and analyses the important impact of viewpoints on Re-ID in detail. Viewpoint-based methods use hard viewpoint label of image directly and utilize them to help features learning. DVAML \cite{ijcai2018-86} tries to learn two different feature sub-spaces for image pairs with similar and dissimilar viewpoints but gets little improvement.

\begin{figure*}[t]
\centering
\includegraphics[width=0.8\textwidth]{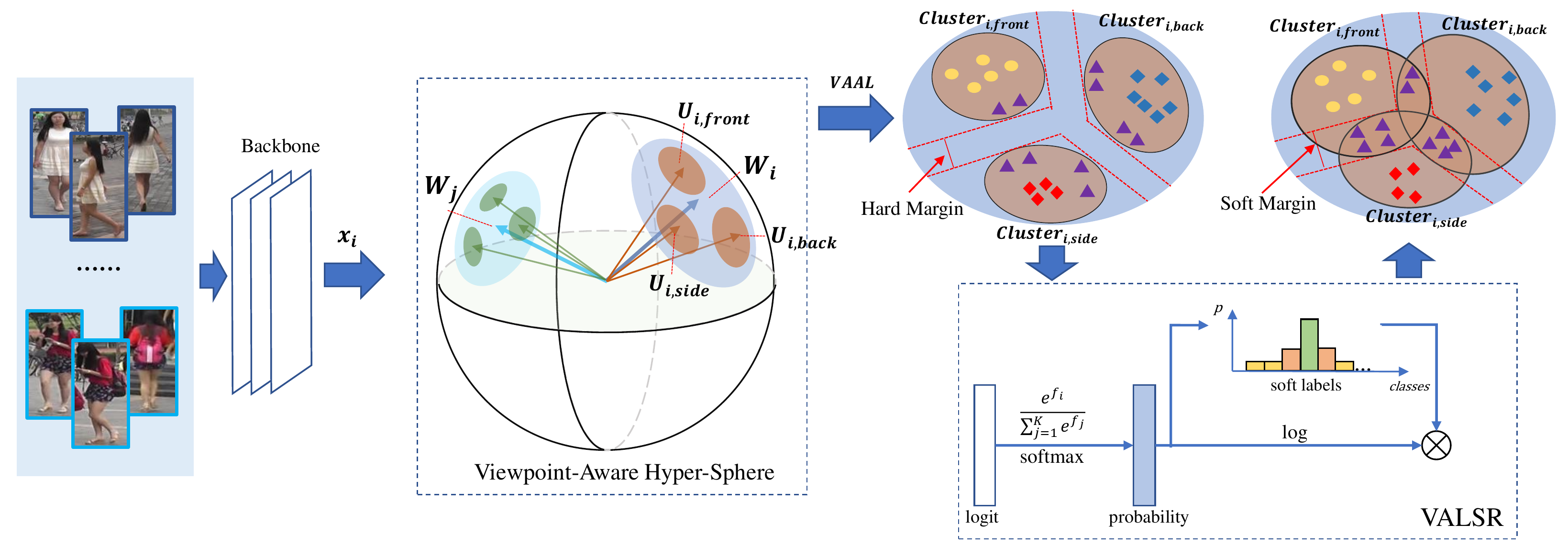} 
\caption{Overview of the proposed VA-reID method. Feature $x_i$ is extracted by backbone network. 
The proposed viewpoint-aware angular loss projects features onto a hyper-sphere to form the identity-level cluster (light green and blue circle) and the viewpoint-level cluster (dark green and brown circle). Furthermore, our Adaptive Label Smoothing Regularization eliminates the hard margin between clusters by introducing adaptive soft labels. }
\label{fig:framework}
\end{figure*}

\noindent \textbf{Loss Function for Identification}. 
Metric learning targets at learning a metric space in which the samples from different classes are far away while the samples from the same classes are compact. The popular loss functions include the softmax-based loss, the contrastive loss and the triplet loss.
In face recognition field, amount of representative softmax-based methods \cite{Salimans2016WeightNA,Liu2017SphereFaceDH,Ranjan2018CrystalLA,Deng_2019_CVPR} have been proposed.These improvements are mainly concentrated on two aspects: normalization and margin. The former is applied for features and weight of fully connected layer. As early as in \cite{Salimans2016WeightNA}, the advantages of weight normalization, such as reducing computational complexity and making the network converge faster, have been demonstrated. Crystal loss \cite{Ranjan2018CrystalLA} proposes feature normalization and verified the effectiveness. Feature normalization is beneficial to increase angular discrimination in feature space direction. The combination of feature normalization and weight normalization can achieve better results. Recent softmax-based work mainly focused on the latter. The concept of angular margin is firstly introduced to metric space in L-Softmax \cite{Liu2016LargeMarginSL}. Later works such as SphereFace \cite{Liu2017SphereFaceDH}, ArcFace \cite{Deng_2019_CVPR} continue to make some improvements to obtain better performance.

\section{Methodology}
\subsection{Problem Formulation and Overview}
Given a query image, the target of Re-ID is to get a ranking list of images from gallery set across non-overlapping camera views. Define an image  $I_i = (x_i, y_i, v_i) $ where $x_i$ denotes the feature extracted by a deep model of the $i$-th image, $y_i$ is the identity label and $v_i$ is the viewpoint label. Notice that each image $x_i$ has only one viewpoint label $v_i\in \{front, side, back\}$. Given a training set $D = \{(x_i, y_i, v_i) \}$ with $N$ images, the deep feature $x_i$ is a global ReID feature extracted by a CNN backbone (\textit{e.g.}, ResNet, DenseNet) denoted as $F(\Theta; I_i )$ where $\Theta$ denotes the parameters of CNN.

Now we briefly introduce the pipeline of the proposed Viewpoint-Aware ReID method (\textbf{VA-reID}). As shown in Figure \ref{fig:framework}, a CNN backbone extracts a global feature $x_i$ for each image $I_i$. We introduce two types of losses: the identity angular loss $L_y$ and the viewpoint-aware angular loss $L_v$. Integrating two angular-related losses into a unified framework of learning can build a two-level distribution for feature $x_i$ on a unified hyper-sphere, including the identity-level distribution and the viewpoint-level distribution. For the identity-level distribution, features with the same identity can be assembled to form identity clusters by $L_y$ (\textit{i.e.}, the large light green and the light blue circles in Figure \ref{fig:framework}). For the viewpoint-level, in each identity cluster, we pull the features of the same viewpoint close to form viewpoint clusters (\textit{i.e}, the small-dark green and the brown circle inside the large circle) by angular loss $L_v$ and center regularization term $L_R$. As a result, the overall loss of our proposed method is:
\begin{equation}
    L_{va} =  L_{y} +  L_v + \beta L_{R}.
\end{equation}
Furthermore, as shown in the bottom-right rectangle in Figure \ref{fig:framework}, at the viewpoint-level, a novel viewpoint-aware adaptive LSR regularization item is used in $L_{v}$ to eliminate the hard margin between viewpoint clusters. At the identity-level, a small adaptive label smoothing regularization is explored to effectively increase the generalization ability of our model.

\subsection{VA-reID for Person Re-Identification}
Softmax loss is widely used for Re-ID feature learning. Let $K$ denote the number of identities, $x$ denote the deep learning ReID feature of a image, and $W_y$ denote the $y$-th column of the weights $W \in \mathbb{R}^{d * K}$. As a result,  the prediction probability of $I$ belongs the identity $y$ is:
\begin{equation}
    q(y) = \frac{e^{W^T_{y}x}}{\sum_{j=1}^{K}e^{W^T_jx}}.
\end{equation}

We remove the bias and normalize the feature $x$ and the weight $W$ as advised in arcface loss \cite{Deng_2019_CVPR}. Generally, weight normalization helps to improve class imbalance, and feature normalization is instrumental in generalization of metric space. Therefore, all the features can be projected onto a hypershere with the same length, and the probability of deep feature $x$ belonging to identity $y$  is equivalent to the cosine distance between $x_i$ and $W_{y}$:
\begin{equation}
    q(y) =  \frac{e^{s\,cos(\theta_{y}-m)}}{e^{s\,cos(\theta_{y}-m)}+\sum_{j=1,j\neq y}^{K}e^{s\,cos(\theta_{j})}}
\label{eq_arcface_id}
\end{equation}
where $\theta_{y}$ denotes the angle between the feature vector $x$ and the $y$-th column of  $W_y$: $cos(\theta_{y}) = \frac{W^T_{y} \cdot  x}{||W_{y}|| \cdot ||x||}$. $m$ is a margin that is used to improve the discriminative ability of classification, and $s$ is the scale factor used to promote convergence.

Based on previous experience \cite{Liu2017SphereFaceDH,Deng_2019_CVPR}, the angular loss helps to model a better discriminative distribution of identities. Analogously, we extend the identity-level angular loss to the proposed viewpoint-aware loss (VA-reID). From Eq.(\ref{eq_arcface_id}), we observe that the $y$-th column of the weight $W_{y}$ can be viewed as the center of identity $y$. To get a higher probability of image $I$ belonging to $y$, we need to pull the feature vector $x_i$ closer to center $W_{y}$. To model the distribution of different viewpoints, each identity class is further classified into $V$ subclass, corresponding to $V$ subclasses corresponding to $V$ viewpoints (\textit{i.e.}, front viewpoint, side viewpoint and back viewpoint. $V$ denotes the number of viewpoints and is assigned 3 in this paper). We denote the viewpoint centers as $U \in \mathbb{R}^{d * K * V}$. As a result, given a deep feature $x$, we model the probability that $x$ belongs to the identity $y$ and viewpoint $v$ as follows:
\begin{equation}
    r(y, v) =  \frac{e^{s\,cos(\phi(y,v)-m)}}{e^{s\, cos(\phi(y, v)-m)}+\sum_{l=1,l\neq y}^{K}\sum_{o=1,o\neq v}^{V}e^{l\,cos(\phi(l, o))}}
\label{eq_arcface_id}
\end{equation}
where $\phi_{yv}$ denotes the angle between feature $x$ and the center of the $y$-th identity and the $v$-th viewpoint $U_{yv}$: $cos(\phi(y, v)) = \frac{U^T_{yv}\cdot x}{||U_{yv}|| \cdot ||x||}$.

Given a training sample $I_i$ with identity label $y_i$ and viewpoint label $v_i$, the identity classification loss $L_y$ and viewpoints-aware loss $L_v$ are:
\begin{equation}
    L_y = \frac{1}{K}\sum_j^K p_jlog(q(j))
\end{equation}

\begin{equation} 
    L_v = \frac{1}{K}\sum_j^K\sum_k^V t_{j,k}log(r(j, k))
\end{equation}
where $p$ and $t$ is the classification label. The traditional methods use the hard label for features learning, \textit{e.g.},
\begin{equation}
     p_j = \left\{
       \begin{array}{ll}
           1 & \textrm{if $i=y_i$} \\
           0 & \textrm{otherwise} \\
       \end{array}
     \right.
\end{equation}

\begin{equation}
     t_{jk} = \left\{
       \begin{array}{ll}
           1 & \textrm{if $j=y_i$ and $j = v_i$} \\
           0 & \textrm{otherwise} \\
       \end{array}
     \right.
\end{equation}
which ignore the ambiguity problem on viewpoint labelling. To overcome this problem, we introduce the soft label learning methods. Furthermore, to maintain the visual similarity between features from the same person but with different viewpoints,  we propose to the center regularization term to connect identity and viewpoint discrimination. It helps to pull the $V$ viewpoint centers $W_{ij}$ closer to the corresponding identity center and the formula is:
\begin{equation}
    L_{R} = \sum_{k=1}^{K} \sum_{j=1}^{V}\frac{W_{j}^T \cdot U_{jk}}{||W_{j}|| \cdot ||U_{jk}||}.
\end{equation}

\begin{figure}[t]
\centering
\includegraphics[width=0.9\columnwidth]{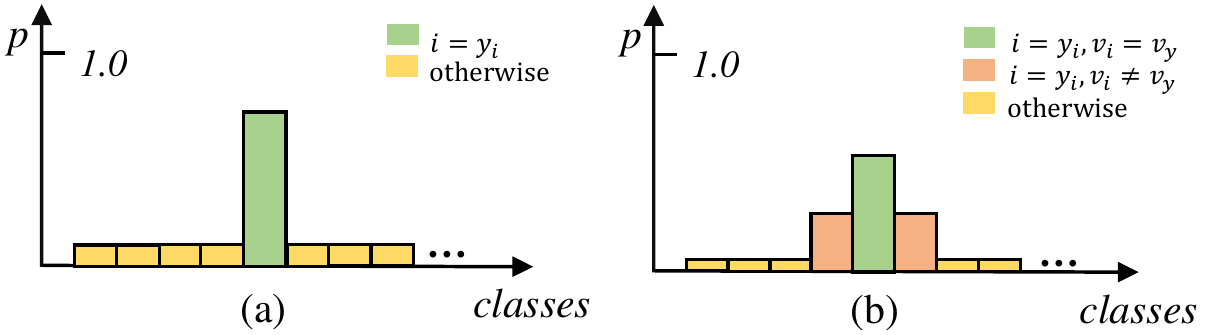} 
\caption{(a) Illustration of adaptive soft label for identity-level learning. (b) Illustration of viewpoint-aware adaptive soft label for viewpoint-level learning .}
\label{fig:prob}
\end{figure}

\noindent \textbf{ - Adaptive Identity Label Learning.}
In this section, we introduce the soft identity label to replace the conventional hard identity label. The soft label will get self-adaption dynamically according to the prediction probability and have better performance against noise data and over-fitting.

{\bf Assumption 1.} \emph {The network tends to prioritize learning simple patterns of real data firstly and then the noise \cite{Arpit2017ACL}.}

As aforementioned, in cross entropy loss, the one-hot encoding is used to be as ground-truth probability distribution and thus the model tends to maximize the expected log-likelihood of one label, which may result in over-fitting and harm the generalization ability of model. Label smoothing regularization(LSR) \cite{Szegedy_2016_CVPR} is proposed to further address the problems. The probability distribution formula of LSR is
\begin{equation}
     p_j = \left\{
       \begin{array}{ll}
           1-\varepsilon & \textrm{if $j=y$} \\
           \displaystyle\frac{\varepsilon}{K-1} & \textrm{otherwise} \\
       \end{array}
     \right.
\end{equation} 
where $\varepsilon$ is a manual value. It replaces the one-hot hard label with the soft label by introduce a small manual parameter $\varepsilon$ to adjust the probability distribution. This encourages the model to lower some confidence on other categories except the ground-truth. However, notice that $\varepsilon$ is a fixed value, which results in the same expected probability of ground-truth category for every input sample and so do other categories. Actually input samples have different logits and applying the same expected probability for them is unsuitable. Considering a case that there exist noises in training set, an image $I$ with the true identity label $y_1$ is annotated as $y_2$ wrongly. According the assumption \cite{Arpit2017ACL}, the network tends to learn positive samples firstly and then noise samples. And usually noise samples have smaller logits of label category than that of positive samples.
\begin{equation}
     p_j = \left\{
       \begin{array}{ll}
           1-\alpha (1-q(j)) & \textrm{if $j=y$} \\
           \displaystyle\frac{\alpha (1-q(j))}{K-1} & \textrm{otherwise} \\
       \end{array}
     \right.
\label{eq_adaptive_lsr}
\end{equation}

We develop a new smoothing parameter $\varepsilon = \alpha (1-q(j)$ , which is related to the prediction probability of network. $\alpha$ is a multiplicative scaling coefficient. When the output probability $q(j)$ is large, we will get a small $\varepsilon$ and get a large confidence on this category. In contrast, when the output probability $q(j)$ is small, we will get large $\varepsilon$ and a large confidence on other categories. The adaptive label smoothing regularization have better performance against noise data and over-fitting. An illustration of the adaptive soft identity label is in Figure \ref{fig:prob}(a). We apply the adaptive-LSR (ALSR) into the identity-level by subsituting Eq.(\ref{eq_adaptive_lsr}) into $L_y$. 

\noindent \textbf{ - Viewpoint-Aware Adaptive Label Learning.}

{\bf Assumption 2.} \emph {The viewpoint of person in reality is a continuous value rather than the hard one. }

Based on this assumption, we extend the ALSR to a viewpoint-aware angular loss, \textit{i.e.}, Viewpoint-Aware ALSR (VALSR). As mentioned earlier, our proposed viewpoint-aware angular loss models two levels of distribution, \textit{i.e.}, the identity-level distribution (identity clusters) and the viewpoint-level distribution (viewpoint clusters). We split every identity label into three sub-categories according to the viewpoints (front, side, back) and thus each image is classified into $3K$ viewpoint-aware categories. We argue that, when assigning the soft label for viewpoint-aware angular loss, the degree of regularization will vary according to the level of cluster that the label belongs to. If current unassigned viewpoint-aware soft label is in the same identity cluster with the ground-truth label's cluster, a stronger relaxation (higher probability) should be assigned as shown in Figure \ref{fig:prob}(b), because images in this cluster will have the strong correlation and the visual similarity with the ground-truth. On the opposite, as shown in Figure \ref{fig:prob}(b), if current soft label is in the different identity cluster with the ground-truth, a weaker relaxation should be assigned. As a result, given a training sample $I_i$ with identity label $y_i$ and viewpoint label $v_i$, the soft label for viewpoint-aware ALSR is
\begin{equation}
t_{jk} = \left\{ \begin{array}{ll}
1-\varepsilon_1-\varepsilon_2 & \textrm{if $j=y_i, k=v_i$}\\
\displaystyle\frac{\varepsilon_2}{2} & \textrm{if $j=y_i, k \neq v_i$}\\
\displaystyle\frac{\varepsilon_1}{K-3} & \textrm{otherwise} 
\end{array} \right.
\label{eq_va_adaptive_lsr}
\end{equation}
where $\varepsilon_1=\alpha(1- \sum_{v=1}^{3}r(y_i, v))$ and $\varepsilon_2 = \alpha(1-r(y_i, v_i))$. We apply the viewpoint adaptive-LSR into angular loss by subsituting Eq.(\ref{eq_va_adaptive_lsr}) into $L_v$.

\subsection{Joint Global and Local Features}
Furthermore, we expect to build a model for both global and local features extraction. VA-reID method have excellent performance of feature extraction, especially for images with various viewpoints. Interestingly, we observe that the viewpoint label are more suitable for the whole body rather than body parts due to huge similarity of body parts with different viewpoint (\textit{e.g.}, lower body, leg). Thus, we mainly apply our VA-reID method to global feature.

\begin{figure}[!t]
\centering
\includegraphics[width=0.9\columnwidth]{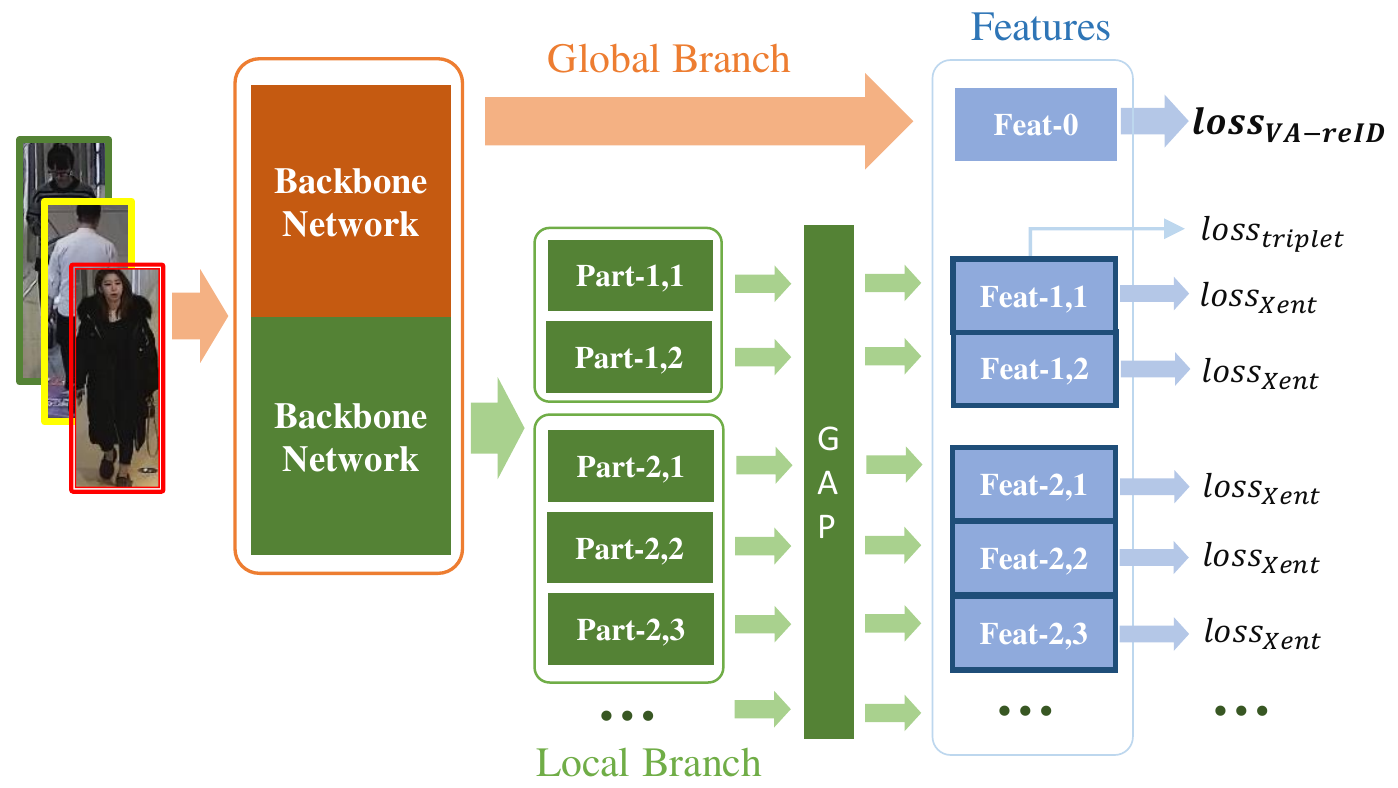}  
\caption{Jointing global and local features. the VA-reID method is used to extract global features while multi-stripes structure is for local features. Xent: cross entropy.}
\label{fig:joint}
\end{figure}

In order to extract local features effectively, we choose the classical multistripe pyramid structure, which is similar to \cite{Sun_2018_ECCV,Zheng_2019_CVPR} for local branches, as shown in Figure \ref{fig:joint}. Jointing the global and local features could effectively boost the performance of the Re-ID model.

\section{Experiments}
\subsection{Datasets and Evaluation Metrics.}
We annotate the viewpoint label\footnote{Available at https://github.com/zzhsysu/VA-ReID} of two widely used benchmarks including Market-1501 and DukeMTMC-reID. Viewpoints are divided into three categories: \textit{front, side, back}. We evaluate our model in the two datasets. \textbf{Notice that we use the viewpoint label only for training}. During the test stage, we don't use any viewpoint label.

\noindent \textbf{Market-1501} 
dataset contains 32,668 person images of 1,501 identities captured by six cameras. Training set is composed of 12,936 images of 751 identities while testing data is composed of the other images of 750 identities. In addition, 2,793 distractors also exist in testing data.

\noindent \textbf{DukeMTMC-reID} 
dataset contains 36,411 person images of 1,404 identities captured by eight cameras. They are randomly divided, with 702 identities as the training set and the remaining 702 identities as the testing set. In the testing set, For each identity in each camera, one image is picked for the query set while the rest remain for the gallery set.

\noindent \textbf{Evaluation Metrics.}
Two widely used evaluation metrics including mean average precision (mAP) and matching accuracy (Rank-1/Rank-5) are adopted in our experiments.

\begin{table*}[!htbp]
\centering
\small
\caption{Performance (\%) comparisons to the state-of-the-art results on Market-1501 and DukeMTMC-reID. Our proposed VA-reID model outperforms the state-of-the-art methods. } 
\begin{tabular}{cllccccc}
\hline
\multirow{2}*{Category} & \multirow{2}*{Method} & \multicolumn{3}{c}{Market-1501} & \multicolumn{3}{c}{DukeMTMC-reID} \\
\cline{3-8}
& & {mAP}&{Rank-1}&{Rank-5}&{mAP}&{Rank-1}&{Rank-5}\\
\hline 
\multirow{3}*{stripe based}& PCB \cite{Sun_2018_ECCV} &  77.4 & 92.3 & 97.2 & 66.1 & 81.7 & 89.7 \\
& PCB+RPP \cite{Sun_2018_ECCV} & 81.6 & 93.8 & 97.5 & 69.2 & 83.3 & 90.5 \\
& MGN \cite{wang2018learning}  & 86.9 & 95.7 & - & 78.4 & 88.7 & - \\ 
\hline 
\multirow{2}*{attention based} & HA-CNN \cite{Li_2018_CVPR} & 75.7 & 91.2 & - & 63.8 & 80.5 & - \\
& ABD-Net \cite{chen2019abd} & 88.28 & 95.60 & - & 78.59 & 89.00 & - \\
\hline
\multirow{2}*{human parsing}& SPReID \cite{Kalayeh_2018_CVPR} & 83.36 & 93.68 & 97.57 & 73.34 & 85.95 & 92.95 \\
& DSA-reID \cite{Zhang_2019_CVPR} & 87.6 & 95.7 & 98.4 & 74.3 & 86.2 & \\
\hline
\multirow{4}*{metric learning } & Pyramid \cite{Zheng_2019_CVPR}  & 88.2 & 95.7 & 98.4 & 79.0 & 89.0 & - \\
& SRB(ResNet50) \cite{Luo_2019_CVPR_Workshops}  & 85.9 & 94.5 & - & 76.4 & 86.4 & - \\
& SRB(SeResNext101) \cite{Luo_2019_CVPR_Workshops} & 88.0 & 95.0 & - & 79.0 & 88.4 & - \\ 
& HPM \cite{fu2019horizontal}  & 82.7 & 94.2 & 97.5 & 74.3 & 86.6 & - \\
\hline
\multirow{5}*{pose/view related} & OSCNN \cite{chen2018person} &  73.5 & 83.9 & - & - & - & - \\ 
& PDC \cite{Su_2017_ICCV} & 63.41 & 84.14 & - & - & - & - \\     
& PN-GAN \cite{Qian_2018_ECCV} & 72.58 & 89.43 & - & 53.20 & 73.58 & - \\ 
& PIE \cite{zheng2019pose} & 69.25 & 87.33 & 95.56 & 64.09 & 80.84 & 88.30 \\
& PGR \cite{li2019pose} & 77.21 & 93.87 & 97.74 & 65.98 & 83.63 & 91.66 \\ 
\hline
\multirow{2}*{\textbf{This work}} & \textbf{Ours} & \textbf{91.70} & \textbf{96.23} & \textbf{98.69} & \textbf{84.51} & \textbf{91.61} & \textbf{96.23}\\
& \textbf{Ours+reranking} & \textbf{95.43} & \textbf{96.79} & \textbf{98.31} & \textbf{91.82} & \textbf{93.85} & \textbf{96.50} \\
\hline
\end{tabular}
\label{tab:tab_compare}
\end{table*}

\subsection{Implementation Details.}
We resize images to $384 \times 128$ as in many re-ID systems. In training stage, we set batch size to be 64 by sampling 16 identities and 4 images per identity. The SeResnext model with the pretrained parameters on ImageNet is considered as the backbone network. Some common data augmentation strategies include horizontal flipping, random cropping, padding, random erasing (with a probability of 0.5) are used. We adopt Adam optimizer to train our model and set weight decay $5 \times 10^{-4}$. The total number of epoch is 200 and the epoch milestones are ${50,100,160}$. The learning rate is initialized to $3.5 \times 10^{-4}$ and is decayed by a factor of 0.1 when the epoch get the milestones. At the beginning, we warm up the models for 10 epochs and the learning rate grows linearly from $3.5 \times 10^{-5}$ to $3.5 \times 10^{-4}$. The parameters in the loss function are set as follows: $\beta=1$, $\alpha=0.2$. 

\subsection{Comparison to the State-of-the-art.}
We evaluate our proposed VA-reID model with the state-of-the-art methods based on deep learning. These methods include: 1) the stripe based methods PCB, MGN; 2) the metric learning related methods SRB,Pyramid,HPM; 3) the human semantic parsing based methods SPReID,DSA-reID; 4) the attention mechanisms based methods HA-CNN,ABD-Net; 5) the pose/view related methods OSCNN, PDC, PN-GAN, PIE, PGR. We show the results in Table \ref{tab:tab_compare} and we can observe that our model achieves state-of-the-art.

\noindent \textbf{ - Comparison to the Pose/View Related Methods.}
Our model outperforms the pose/view-related methods. Without reranking, our model achieves an improvement over the best pose/view-related method PGR by 14.49\%/2.36\% on mAP/Rank-1 metrics in Market-1501 and by 18.53\%/7.98\% on mAP/Rank-1 metrics in DukeMTMC-reID.

\noindent \textbf{ - Comparison to the Metric Learning Related Methods.}
Our model outperforms the metric learning related methods. Without reranking, our model achieves an improvement over the second best method SRB by 3.70\%/1.23\% on mAP/Rank-1 metrics in Market-1501 and by 5.51\%/3.21\% on mAP/Rank-1 metrics in DukeMTMC-reID.

\noindent \textbf{ - Comparison to Other Methods.}
Our model outperforms the stripe based, the human semantic parsing based and the attention mechanisms based methods. Comparison to the recent state-of-the-art method ABD-net, our model achieves an improvement by 3.42\%/0.63\% on mAP/Rank-1 metrics in Market-1501 and by 5.92\%/2.61\% on mAP/Rank-1 metrics in DukeMTMC-reID without reranking.

\subsection{Ablation Study.}
We perform comprehensive ablation study to demonstrate: 1) the effectiveness of the adaptive label smoothing; 2) the effectiveness of the viewpoint-aware adaptive label smoothing; 3) the effectiveness of center regularization; 4) the effectiveness of jointing global and local features. Notice that the viewpoint-aware loss function $L_{va}$ is for global branch. Loss of VA-reID is $L_{av}=L_y$+${L_{v}}+\beta {L_R}$, where $L_y$ is adaptive label smoothing loss item, $L_v$ is viewpoint-aware adaptive label smoothing loss item and $L_{R}$ is center regularization. \textbf{We use the model trained only with $L_y$ as the baseline}, and set parameter $\beta$=0.1 and  $\alpha$=0.2. The performance (\%) comparisons of different modules on Market-1501 and DukeMTMC-reID datasets are shown in Table \ref{tab:ablation}.

\noindent \textbf{ - Effectiveness of Adaptive Label Smoothing.}
Comparing results of cross entropy loss (Xent) and label smoothing loss (LSR), we can observe that using label smoothing gets better performance. We use adaptive label smoothing (ALSR) as basic loss and it achieves an improvement over label smoothing by 0.25\%/0.35\% on mAP/Rank-1 metrics in Market-1501 and by 0.52\%/0.45\% on mAP/Rank-1 metrics in DukeMTMC-reID. This is because ALSR replaces the one-hot hard label with the adaptive soft label for identity classification. The adaptive soft label actually helps to learn discriminative features while ignore the negative impact of noises. This comparison demonstrates the effectiveness of adaptive label smoothing.

\noindent \textbf{ - Effectiveness of Viewpoint-Aware Adaptive Label Smoothing.}
Comparing results to the baseline model, we observe that combining the viewpoint-aware adaptive label smoothing(VALSR) and the adaptive label smoothing(ALSR) can achieve an improvement by 1.70\%/0.67\% on mAP/Rank-1 metrics in Market-1501 and by 0.87\%/0.99\% on mAP/Rank-1 metrics in DukeMTMC-reID. This is because VALSR uses the viewpoint-aware adaptive soft label for identity-viewpoint classification. For each identity, the viewpoint-aware adaptive soft label helps to learn the embedding of viewpoint-related compact features. This comparison demonstrates the effectiveness of viewpoint-aware adaptive label smoothing. 

\begin{table}[!t]
\centering
\small
\caption{Performance (\%) comparisons of different modules on Market-1501 and DukeMTMC-reID datasets. RR: using reranking. Xent: cross entropy loss.} 
\label{tab:ablation}
\begin{tabular}{lp{0.6cm}<{\centering}cp{0.6cm}<{\centering}c}
\hline
\multirow{2}*{Method}  & \multicolumn{2}{c}{Market-1501} & \multicolumn{2}{c}{DukeMTMC-reID} \\
\cline{2-5}
 &{mAP}&{Rank-1}&{mAP}&{Rank-1}\\
\hline
Xent & 86.30 & 94.31 & 76.70 & 86.94 \\
LSR & 86.72 & 94.35 & 77.47 & 87.43 \\
$L_y$(baseline) & 86.97 & 94.7 & 77.99 & 87.39 \\
$L_y$+${L_v}$ & 88.67 & 95.37 & 78.86 & 88.38 \\
$L_y$+${L_{R}}$ & 88.25 & 95.25 & 78.25 & 87.84 \\
VA-reID & 89.97 & 95.87 & 81.48 & 91.11\\
VA-reID+RR & 95.09 & 96.32 & 90.66 & 92.46 \\
\hline
VA-reID+local & 91.70 & 96.23 & 84.51 & 91.61 \\
VA-reID+local+RR & 95.43 & 96.79 & 91.82 & 93.85 \\
\hline
\end{tabular}
\end{table}

\noindent \textbf{ - Effectiveness of Center Regularization.}
Comparing results of the baseline model and baseline with ${L_{R}}$, we can observe that adding an center regularization to the baseline model can get a slight improvement. Comparing results of $L_y$+${L_{R}}$ and VA-reID, we can observe an significant improvement by 3.00\%/1.17\% on mAP/Rank-1 metrics in Market-1501 and by 3.45\%/3.27\% on mAP/Rank-1 metrics in DukeMTMC-reID. This is because the center regularization and the viewpoint-aware adaptive soft label can complement each other. For each identity, viewpoint-aware adaptive soft label generates feature clusters of viewpoints while the center regularization makes centers of these clusters closer. This comparison demonstrates the effectiveness of center regularization.

\noindent \textbf{ - Effectiveness of Jointing Global and Local Features.}
Comparing results of the VA-reID model, without reranking, jointing VA-reID and local feature can get a significant improvement by 1.73\%/0.36\% on mAP/Rank-1 metrics in Market-1501 and by 3.03\%/0.50\% on mAP/Rank-1 metrics in DukeMTMC-reID. VA-reID uses viewpoint-aware loss only for learning discriminative global features while the local branch adopts multi-strips structure to learning fine-grained features. Adding local features helps to improve performance further. This comparison demonstrates the effectiveness of jointing global and local features.

\subsection{Further Evaluations.}
Following evaluations and analysis are made to further verify the effectiveness of our methods including: (1) the influence of the noisy viewpoint label; (2) visualization of the influence of viewpoints variance to the retrival; (3) the effect of hyperparameters; (4) viewpoint based retrieval and complexity analysis (see supplementary material\footnote{Available at https://github.com/zzhsysu/VA-ReID}).

\noindent \textbf{ - Influence of the Noisy Viewpoint Label.}
We exam the influence of the noises in the viewpoint labels on the Re-ID performance. 
As show in Table \ref{tab:table_noise_label} three different types of viewpoints labels and the performance comparison using different viewpoint labels:

1) \textbf{VA-reID with P} uses viewpoints labels predicted by a simple Resnet-50 classifier trained on 600 images of extra viewpoints datasets. 2) \textbf{VA-reID with PC} uses viewpoints labels predicted by clustering pose information generated by Open-Pose. 3) \textbf{VA-reID with GT} directly uses ground-truth labels annotated by human. The accuracy of viewpoints classification is 78.7\% and the accuracy of pose clustering is 44.57\%.

From Table \ref{tab:table_noise_label}, we observe that the amount of noises in the viewpoints labels have very little influence on our method. VA-reID with P achieves very similar performance to VA-reID with GT. What's more, even with the viewpoints prediction accuracy of  only 44.57\%, the performance of VA-reID with PC does not drop much, our method still achieves state-of-art performance on DukeMTMC-reID. This experiments verify that our proposed VA-reID method is robust to viewpoints label noises and a simple classifier or clustering method is good enough to get the viewpoints information. 


\begin{table}[!h]
\centering
\small
\caption{Performance (\%) comparisons for different generation methods of the viewpoint label. \textbf{P}: using predictive viewpoint labels. \textbf{PC}: using pose-based clustering viewpoint labels. \textbf{GT}: using ground-truth viewpoint labels.} 
\begin{tabular}{ccccc}
\hline
\multirow{2}*{Method}  &  \multicolumn{2}{c}{DukeMTMC-reID} \\
\cline{2-4}
 &{mAP}&{Rank-1}& {Rank-5}\\
\hline
PN-GAN \cite{Qian_2018_ECCV} & 53.20 & 73.58 & - \\
\hline
VA-reID with P & 81.05 & 90.75 & 95.65 \\ 
VA-reID with PC & 80.69 & 90.35 & 95.83 \\ 
VA-reID with GT & 81.48 & 91.11 & 95.38 \\ 
\hline
\end{tabular}
\label{tab:table_noise_label}
\end{table}

\begin{figure}[!t]
\centering
\includegraphics[width=0.85\columnwidth]{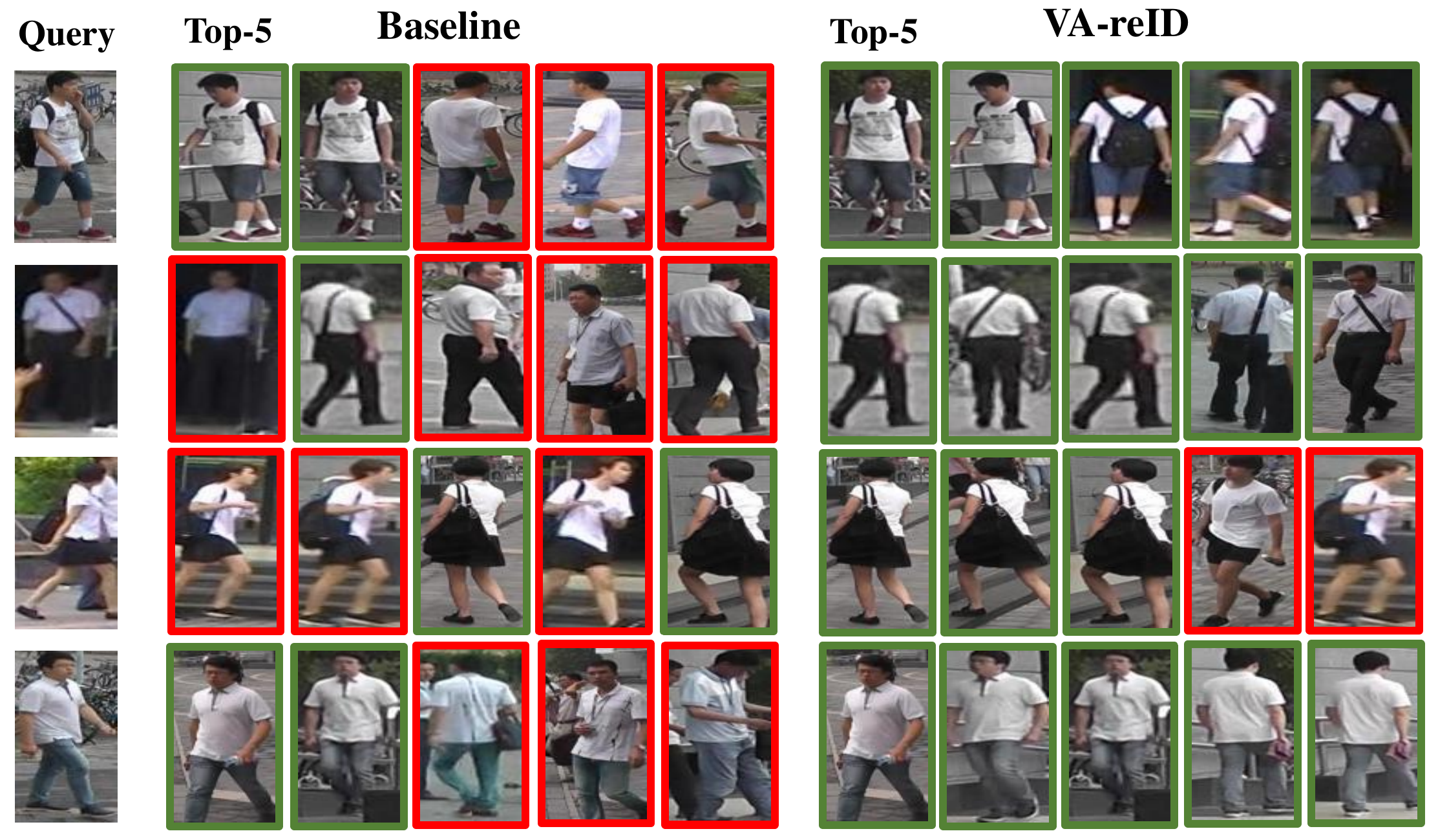} 
\caption{Visual results of the baseline method and the VA-reID method. The red box represents the wrong result while the green box represents the correct result.} 
\label{fig:visdom}
\end{figure}

\noindent \textbf{ - Visualization of Results.}
Figure \ref{fig:visdom} shows examples of retrieval results by baseline method and VA-reID. We observe that, in the case of high viewpoints variance between query and gallery, our method has a much higher retrieval precision compared to baseline. Take the result in the first row as an example, for a query image in front viewpoint, baseline method only correctly retrieves two images with the same front viewpoint, while VA-reID is able to successfully retrieve images from all three viewpoints. Similar to the first row, the results in other rows further demonstrate excellent performance on cross-viewpoints images retrieval. 

\begin{figure}[!h]
\centering
\includegraphics[width=0.85\columnwidth]{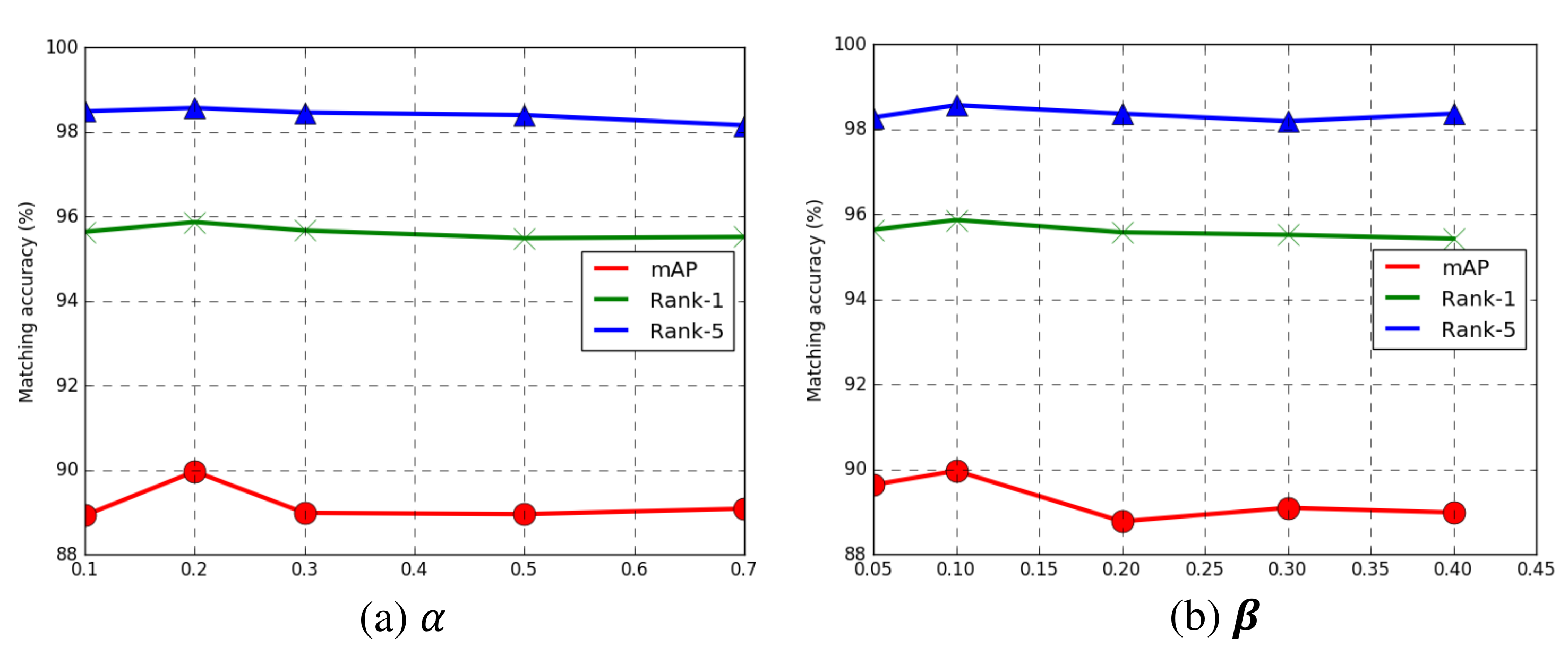} 
\caption{Performance of VA-reID method on Market-1501 with different hyperparameters. In figure (a), we fix ${\beta=0.1}$. In figure (b), we fix ${\alpha=0.2}$.}
\label{fig:ablation}
\end{figure}

\noindent \textbf{ - Hyperparameters Analysis.}
Figure \ref{fig:ablation} shows how the parameter $\alpha$ and $\beta$ affect the performance in Market-1501 dataset. The performace of our method is stable within a wide range for both parameters. It also presents the excellence of the adaptive label smoothing method.

\section{Conclusion}
This study proposes a novel method to address the viewpoint variation problem in Re-ID. Overall, we make two contributions to the community. Firstly, we propose the viewpoint-aware angular loss to learn the embedding of viewpoint-aware feature in a unified hyper-sphere, which effectively model the feature distribution on both identity-level and viewpoints-level. Secondly, we propose a novel viewpoint aware adaptive label smoothing method to relax the hard margin caused with adaptive soft labels. Experiments show the effectiveness our method. 

\section{Acknowledgement}
This work was supported partially by the National Key Research and Development Program of China (2016YFB1001002), NSFC(61522115,U1811461), Guangdong Province Science and Technology Innovation Leading Talents (2016TX03X157), and Guangzhou Research Project (201902010037).

\bibliographystyle{aaai}
\bibliography{aaai}

\begin{thebibliography}{}

\bibitem[\protect\citeauthoryear{Arpit \bgroup et al\mbox.\egroup
  }{2017}]{Arpit2017ACL}
Arpit, D.; Jastrzebski, S.; Ballas, N.; Krueger, D.; Bengio, E.; Kanwal, M.~S.;
  Maharaj, T.; Fischer, A.; Courville, A.~C.; Bengio, Y.; and Lacoste-Julien,
  S.
\newblock 2017.
\newblock A closer look at memorization in deep networks.
\newblock {\em ArXiv} abs/1706.05394.

\bibitem[\protect\citeauthoryear{Chen \bgroup et al\mbox.\egroup
  }{2018}]{chen2018person}
Chen, Y.; Duffner, S.; Stoian, A.; Dufour, J.-Y.; and Baskurt, A.
\newblock 2018.
\newblock Person re-identification with a body orientation-specific
  convolutional neural network.
\newblock In {\em International Conference on Advanced Concepts for Intelligent
  Vision Systems},  26--37.
\newblock Springer.

\bibitem[\protect\citeauthoryear{Chen \bgroup et al\mbox.\egroup
  }{2019}]{chen2019abd}
Chen, T.; Ding, S.; Xie, J.; Yuan, Y.; Chen, W.; Yang, Y.; Ren, Z.; and Wang,
  Z.
\newblock 2019.
\newblock Abd-net: Attentive but diverse person re-identification.
\newblock {\em arXiv preprint arXiv:1908.01114}.

\bibitem[\protect\citeauthoryear{Chen, Xu, and Deng}{2018}]{ijcai2018-86}
Chen, P.; Xu, X.; and Deng, C.
\newblock 2018.
\newblock Deep view-aware metric learning for person re-identification.
\newblock In {\em Proceedings of the Twenty-Seventh International Joint
  Conference on Artificial Intelligence},  620--626.
\newblock International Joint Conferences on Artificial Intelligence
  Organization.

\bibitem[\protect\citeauthoryear{Deng \bgroup et al\mbox.\egroup
  }{2019}]{Deng_2019_CVPR}
Deng, J.; Guo, J.; Xue, N.; and Zafeiriou, S.
\newblock 2019.
\newblock Arcface: Additive angular margin loss for deep face recognition.
\newblock In {\em CVPR}.

\bibitem[\protect\citeauthoryear{Fu \bgroup et al\mbox.\egroup
  }{2019}]{fu2019horizontal}
Fu, Y.; Wei, Y.; Zhou, Y.; Shi, H.; Huang, G.; Wang, X.; Yao, Z.; and Huang, T.
\newblock 2019.
\newblock Horizontal pyramid matching for person re-identification.
\newblock In {\em AAAI}, volume~33,  8295--8302.

\bibitem[\protect\citeauthoryear{Kalayeh \bgroup et al\mbox.\egroup
  }{2018}]{Kalayeh_2018_CVPR}
Kalayeh, M.~M.; Basaran, E.; Gökmen, M.; Kamasak, M.~E.; and Shah, M.
\newblock 2018.
\newblock Human semantic parsing for person re-identification.
\newblock In {\em CVPR}.

\bibitem[\protect\citeauthoryear{Leng, Ye, and Tian}{2019}]{leng2019survey}
Leng, Q.; Ye, M.; and Tian, Q.
\newblock 2019.
\newblock A survey of open-world person re-identification.
\newblock {\em IEEE Transactions on Circuits and Systems for Video Technology}.

\bibitem[\protect\citeauthoryear{Li \bgroup et al\mbox.\egroup
  }{2019}]{li2019pose}
Li, J.; Zhang, S.; Tian, Q.; Wang, M.; and Gao, W.
\newblock 2019.
\newblock Pose-guided representation learning for person re-identification.
\newblock {\em IEEE transactions on pattern analysis and machine intelligence}.

\bibitem[\protect\citeauthoryear{Li, Zhu, and Gong}{2018}]{Li_2018_CVPR}
Li, W.; Zhu, X.; and Gong, S.
\newblock 2018.
\newblock Harmonious attention network for person re-identification.
\newblock In {\em CVPR}.

\bibitem[\protect\citeauthoryear{Liu \bgroup et al\mbox.\egroup
  }{2016}]{Liu2016LargeMarginSL}
Liu, W.; Wen, Y.; Yu, Z.; and Yang, M.
\newblock 2016.
\newblock Large-margin softmax loss for convolutional neural networks.
\newblock In {\em ICML}.

\bibitem[\protect\citeauthoryear{Liu \bgroup et al\mbox.\egroup
  }{2017}]{Liu2017SphereFaceDH}
Liu, W.; Wen, Y.; Yu, Z.; Li, M.~M.; Raj, B.; and Song, L.
\newblock 2017.
\newblock Sphereface: Deep hypersphere embedding for face recognition.
\newblock {\em CVPR}  6738--6746.

\bibitem[\protect\citeauthoryear{Luo \bgroup et al\mbox.\egroup
  }{2019}]{Luo_2019_CVPR_Workshops}
Luo, H.; Gu, Y.; Liao, X.; Lai, S.; and Jiang, W.
\newblock 2019.
\newblock Bag of tricks and a strong baseline for deep person
  re-identification.
\newblock In {\em CVPRW}.

\bibitem[\protect\citeauthoryear{Qian \bgroup et al\mbox.\egroup
  }{2018}]{Qian_2018_ECCV}
Qian, X.; Fu, Y.; Xiang, T.; Wang, W.; Qiu, J.; Wu, Y.; Jiang, Y.-G.; and Xue,
  X.
\newblock 2018.
\newblock Pose-normalized image generation for person re-identification.
\newblock In {\em ECCV}.

\bibitem[\protect\citeauthoryear{Ranjan \bgroup et al\mbox.\egroup
  }{2018}]{Ranjan2018CrystalLA}
Ranjan, R.; Bansal, A.; Xu, H.; Sankaranarayanan, S.; Chen, J.-C.; Castillo,
  C.~D.; and Chellappa, R.
\newblock 2018.
\newblock Crystal loss and quality pooling for unconstrained face verification
  and recognition.
\newblock {\em ArXiv} abs/1804.01159.

\bibitem[\protect\citeauthoryear{Salimans and
  Kingma}{2016}]{Salimans2016WeightNA}
Salimans, T., and Kingma, D.~P.
\newblock 2016.
\newblock Weight normalization: A simple reparameterization to accelerate
  training of deep neural networks.
\newblock In {\em NIPS}.

\bibitem[\protect\citeauthoryear{Saquib~Sarfraz \bgroup et al\mbox.\egroup
  }{2018}]{Sarfraz_2018_CVPR}
Saquib~Sarfraz, M.; Schumann, A.; Eberle, A.; and Stiefelhagen, R.
\newblock 2018.
\newblock A pose-sensitive embedding for person re-identification with expanded
  cross neighborhood re-ranking.
\newblock In {\em CVPR}.

\bibitem[\protect\citeauthoryear{Su \bgroup et al\mbox.\egroup
  }{2017}]{Su_2017_ICCV}
Su, C.; Li, J.; Zhang, S.; Xing, J.; Gao, W.; and Tian, Q.
\newblock 2017.
\newblock Pose-driven deep convolutional model for person re-identification.
\newblock In {\em CVPR}.

\bibitem[\protect\citeauthoryear{Sun and Zheng}{2019}]{Sun_2019_CVPR}
Sun, X., and Zheng, L.
\newblock 2019.
\newblock Dissecting person re-identification from the viewpoint of viewpoint.
\newblock In {\em CVPR}.

\bibitem[\protect\citeauthoryear{Sun \bgroup et al\mbox.\egroup
  }{2018}]{Sun_2018_ECCV}
Sun, Y.; Zheng, L.; Yang, Y.; Tian, Q.; and Wang, S.
\newblock 2018.
\newblock Beyond part models: Person retrieval with refined part pooling.
\newblock In {\em ECCV}.

\bibitem[\protect\citeauthoryear{Szegedy \bgroup et al\mbox.\egroup
  }{2016}]{Szegedy_2016_CVPR}
Szegedy, C.; Vanhoucke, V.; Ioffe, S.; Shlens, J.; and Wojna, Z.
\newblock 2016.
\newblock Rethinking the inception architecture for computer vision.
\newblock In {\em CVPR}.

\bibitem[\protect\citeauthoryear{Wang \bgroup et al\mbox.\egroup
  }{2018}]{wang2018learning}
Wang, G.; Yuan, Y.; Chen, X.; Li, J.; and Zhou, X.
\newblock 2018.
\newblock Learning discriminative features with multiple granularities for
  person re-identification.
\newblock In {\em ACM MM},  274--282.
\newblock ACM.

\bibitem[\protect\citeauthoryear{Zhang \bgroup et al\mbox.\egroup
  }{2019}]{Zhang_2019_CVPR}
Zhang, Z.; Lan, C.; Zeng, W.; and Chen, Z.
\newblock 2019.
\newblock Densely semantically aligned person re-identification.
\newblock In {\em CVPR}.

\bibitem[\protect\citeauthoryear{Zhao \bgroup et al\mbox.\egroup
  }{2017}]{Zhao_2017_ICCV}
Zhao, L.; Li, X.; Zhuang, Y.; and Wang, J.
\newblock 2017.
\newblock Deeply-learned part-aligned representations for person
  re-identification.
\newblock In {\em The IEEE International Conference on Computer Vision}.

\bibitem[\protect\citeauthoryear{Zheng \bgroup et al\mbox.\egroup
  }{2015}]{zheng2015scalable}
Zheng, L.; Shen, L.; Tian, L.; Wang, S.; Wang, J.; and Tian, Q.
\newblock 2015.
\newblock Scalable person re-identification: A benchmark.
\newblock In {\em The IEEE Conference on Computer Vision and Pattern
  Recognition}.

\bibitem[\protect\citeauthoryear{Zheng \bgroup et al\mbox.\egroup
  }{2019a}]{Zheng_2019_CVPR}
Zheng, F.; Deng, C.; Sun, X.; Jiang, X.; Guo, X.; Yu, Z.; Huang, F.; and Ji, R.
\newblock 2019a.
\newblock Pyramidal person re-identification via multi-loss dynamic training.
\newblock In {\em CVPR}.

\bibitem[\protect\citeauthoryear{Zheng \bgroup et al\mbox.\egroup
  }{2019b}]{zheng2019pose}
Zheng, L.; Huang, Y.; Lu, H.; and Yang, Y.
\newblock 2019b.
\newblock Pose invariant embedding for deep person re-identification.
\newblock {\em IEEE Transactions on Image Processing}.

\bibitem[\protect\citeauthoryear{Zhou and Shao}{2018}]{Zhou_2018_CVPR}
Zhou, Y., and Shao, L.
\newblock 2018.
\newblock Viewpoint-aware attentive multi-view inference for vehicle
  re-identification.
\newblock In {\em CVPR}.

\end{thebibliography}

\end{document}